\documentclass[runningheads]{llncs}
\usepackage[T1]{fontenc}
\usepackage{amsmath}
\usepackage{array}
\usepackage{float}
\usepackage{graphicx}
\usepackage{amssymb}
\usepackage{tikz}
\usepackage{multirow}
\usepackage[pageanchor=false,colorlinks=true,linkcolor=blue,citecolor=blue,urlcolor=blue]{hyperref}
\usetikzlibrary{backgrounds}
\pgfdeclarelayer{background}
\pgfdeclarelayer{lowerbackground}
\pgfsetlayers{lowerbackground,background,main} 
\newcommand{\argmin}[1]{\underset{#1}{\operatorname{arg}\,\operatorname{min}}\;}

\begin{document}
\title{Parameter-free structure-texture image decomposition by unrolling}

\author{Laura Girometti\inst{1} \and
Jean-François Aujol\inst{2} \and
Antoine Guennec \inst{2}
\and 
Yann Traonmilin\inst{2}}
\authorrunning{L.Girometti et al.}
%
\institute{Department of Mathematics, University of Bologna \\\email{laura.girometti2@unibo.it} \and
Institut de Mathématiques de Bordeaux, IMB\\
\email{\mbox{\{jean-francois.aujol, antoine.guennec, yann.traonmilin\}@math.u-bordeaux.fr}}}
\maketitle
\begin{abstract}
In this work, we propose a parameter-free and efficient method to tackle the structure-texture image decomposition problem. In particular, we present a neural network LPR-NET based on the unrolling of the Low Patch Rank model.
On the one hand, this allows us to automatically learn parameters from data, and on the other hand to be computationally faster while obtaining qualitatively similar results compared to traditional iterative model-based methods. Moreover, despite being trained on synthetic images, numerical experiments show the ability of our network to generalize well when applied to natural images.

\end{abstract}
\section{Introduction}
The problem of structure-texture image decomposition has been widely investigated in the last decades and successfully applied to tasks such as detail enhancement, image abstraction, texture transfer, HDR tone mapping \cite{PnP,unsupervised}. In addition, it has also been exploited to solve inverse problems in imaging such as image fusion \cite{FUSION}, inpainting \cite{GEN_MOD} and super-resolution \cite{SMOOTHPLUSSPARSE}.  It aims to find an additive decomposition of an observed image $f$ of the form
\begin{equation}
    f = u+v,
\end{equation}
where $u$ represents the structure component of $f$ containing the geometrical elements identified in edges and homogeneous regions, whereas the texture component $v$ contains repeated patterns of small scale details.
This problem is usually formulated as a constrained minimization problem as follows:
\begin{equation}
    \min_{u,v \in \mathbb{R}^N \; s.t. f = u+v} \mu \mathcal{J}_1(u) + \mathcal{J}_2(v),
    \label{eq:STD_general}
\end{equation}
where $\mathcal{J}_1$ and $\mathcal{J}_2$ are regularization functions that enforce the desired properties on the components $u$ and $v$, and $\mu$ is a positive weighting parameter that plays a crucial role in separating $u$ from $v$, adapting to the image content. While for the structure component the natural choice for $\mathcal{J}_1$ falls on functions promoting gradient-sparsity - Total Variation \cite{ROF}, Non convex Total Variation \cite{Non-convexPHI} - various proposals for $\mathcal{J}_2$ have been deeply investigated:  high-oscillatory pattern promoting functions - G-norm \cite{G-norm,TV-G} - low patch rank penalties that promote similarity between texture patches - \cite{LPR,BNN,localLPR}, a generative texture prior inspired by generative convolutional neural networks \cite{GEN_MOD}, convolutional sparse coding with dictionary learning-based sparsity prior \cite{SPARSECODING,SMOOTHPLUSSPARSE}. 
Another line of research focuses on optimal parameter selection based on separability between components- minimize the cross-correlation between pairs of them \cite{quaternary,struct} - or on adjusting the tuning parameters depending on the right relationship between the gradient sparsity level of $u$ and the rank of the texture patches \cite{localLPR}. Recently,
motivated by progress in machine learning techniques, various data-driven approaches have been presented. The problem of lack of available datasets containing pairs of structure-texture components has been addressed by applying an unsupervised strategy \cite{unsupervised}. The Plug-and-Play framework, where a denoiser is trained to learn a more significant regularizer for the structure $u$ on natural image datasets \cite{PnP} enables the learning of a joint regularizer $R_x(u,v)$ for structure and texture on a generated synthetic  dataset \cite{PnP_joint}.
Among the hybrid techniques that combine model-based methods and deep-learning approaches, algorithm unrolling is a recent promising paradigm to
build efficient and interpretable neural networks. These techniques date back to Gregor et al.’s paper \cite{GregLecun}, where they proposed to improve the computational efficiency of
sparse coding algorithms through end-to-end training. Following their idea, the guiding principle in building unrolled neural networks is to represent each iteration of an iterative algorithm as one layer of the network. Consequently, mimicking model-based methods results in high interpretability of network layers, fewer parameters compared with popular neural networks and, therefore, less training data, as well as better generalization power than generic networks \cite{review}. Recently, it has been applied to different signal and image processing tasks such as image denoising \cite{den_unrolling}, blind image deblurring \cite{deblur} and CT \cite{OktemPD}. \\
This paper proposes a novel strategy to achieve an efficient and automatic structure-texture image decomposition by leveraging the unrolling technique: model and algorithm parameters are learned in a supervised fashion making use of a synthetic generated dataset.
In particular, in Section 2, we propose to unroll an extension of the Low Patch Rank model \cite{LPR} and we present a numerical solution based on the Alternating Direction Method of Multipliers (ADMM). In Section 3, we propose our neural network LPR-NET and we investigate various design choices. Finally, in Section 4, numerical results validate that our proposal outperforms optimally hand-tuned classical variational models and gives a competitive end-to-end alternative to the Plug-and-Play framework where optimization of the learned regularization must be performed.

\section{Proposed structure-texture decomposition model}
In this section, we first describe an extended version of the Low Patch Rank model for structure-texture decomposition and we present an ADMM-based algorithm to solve it. Then, we propose our LPR-NET network based on the unrolling of the proposed ADMM iterative scheme.  
\subsection{Low Patch Rank model for structure-texture decomposition}
Given an image $f$ possibly degraded by a linear operator $\mathcal{M}$ (e.g. deblurring, mask operator), the Low Patch Rank model \cite{LPR} recovers the structure component $u$ and the texture component $v$ by enforcing $u$ to be gradient sparse and $v$ to be of low patch rank. Specifically, they consider the classical Total Variation as regularization function $\mathcal{J}_1$ and the nuclear norm of texture patches as $\mathcal{J}_2$. We propose to extend this model by considering, instead of the $\ell_1$-norm, a non-convex non-smooth sparsity promoting function $\phi(\cdot;a)$ parametrized by $a\geq 0$. The resulting variational problem reads as:
\begin{equation}
    \min_{u \in \mathbb{R}^N, v \in \mathbb{R}^N  s.t. f = \mathcal{M}(u + v)} \mu \sum_{i=1}^N \phi(||(\mathrm{D} u)_i ||_{2}; a) +  ||\mathcal{P} v ||_*,
\label{eq:TV-LPR}
\end{equation}
where $\mu$ is a positive balancing parameter, $(\mathrm{D} u)_i$ represents the discrete gradient of $u$ at pixel $i$ and the nuclear norm  $||\cdot||_*$ is chosen as a convex surrogate of the rank function. The operator $\mathcal{P}$ represents the patch operator characterized by the size $p$ of the $p \times p$ patches and by the overlap $o$ between two adjacent patches as shown in Figure 1: 
\begin{equation}
   \mathcal{P} : \mathbb{R}^{n \times m} \to \mathbb{R}^{p^2\times N_{patches}}
    \nonumber
\end{equation}
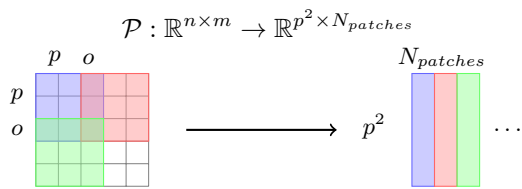
\begin{figure}[H]
\centering
\vspace{-10mm}
\begin{tikzpicture}
\def\a{0}
    \draw[step=0.3cm,gray,very thin] (0+\a,0) grid (1.5+\a,1.5);
    \draw[color = blue!60!white,fill = blue!40!white, fill opacity = 0.5] (0+\a,1.5) rectangle (0.9+\a,0.6);
    \draw[color = red!60!white,fill = red!40!white, fill opacity = 0.5] (0.6+\a,1.5) rectangle (1.5+\a,0.6);
    \draw[color = green!60!white,fill = green!40!white, fill opacity = 0.5] (0+\a,0.9) rectangle (0.9+\a,0);
      \node at (-0.25+\a,1.2) {$p$};
       \node at (-0.25+\a,0.75) {$o$};
     \node at (0.25+\a,1.7) {$p$};
     \node at (0.7+\a,1.7) {$o$};
     \draw[thick,->] (2+\a,0.75) -- (4+\a,0.75);
     \node at (4.5+\a, 0.8) {$p^2$};
     \node at (5.4 +\a,1.7) {$N_{patches}$};
     \draw[color = blue!60!white,fill = blue!40!white, fill opacity = 0.5] (5+\a,0) rectangle (5.3+\a,1.5);
     \draw[color = red!60!white, fill = red!40!white, fill opacity = 0.5] (5.3+\a,0) rectangle (5.6+\a,1.5);
      \draw[color = green!60!white, fill = green!40!white, fill opacity = 0.5] (5.6+\a,0) rectangle (5.9+\a,1.5);
     \node at (6.3+\a,0.75) {$\dots$};
\end{tikzpicture}
\caption{ Graphical action of the patch operator $\mathcal{P}$.}
\end{figure}
The considered non-convex penalty function $\phi(\cdot;a)$ is the minimax concave penalty (MCP) \cite{MCP} defined as 
\begin{equation}
    \phi(t;a) = \left\{
    \begin{array}{l}
         |t| - \frac{a}{2} t^2 \;  \; \mathrm{if} \; |t| \leq \frac{1}{a}  
         \vspace{0.1cm} \\
         \frac{1}{2 a}   \; \quad \qquad \mathrm{if} \; |t| \geq \frac{1}{a}
    \end{array}.
    \right.
\end{equation}
Note that as $a \to 0 $, we recover the absolute value function $|t|$ as a special case of the penalty function $\phi(\cdot;a)$ thus the classical Total Variation regularizer in \eqref{eq:TV-LPR}. The MCP function was first introduced in \cite{MCP} in the context of unbiased variable
selection to overstep the bias introduced by the $\ell_1$ penalty on large coefficients. Therefore, we aim to less penalize high gradients in the structure component $u$.


\subsection{ADMM-based algorithm for LPR}
To numerically solve the constrained minimization problem in \eqref{eq:TV-LPR}, we equivalently reformulate the problem introducing two auxiliary variables $t := \mathrm{D}u \in \mathbb{R}^{2N}$ and $s:= v \in \mathbb{R}^N$, which represent the discrete gradient of $u$ and the texture $v$ respectively. Problem~\eqref{eq:TV-LPR} is rewritten as
\begin{equation}
     \min_{\small{\begin{array}{c}
       t \in \mathbb{R}^{2N}, u, v, s \in \mathbb{R}^N     \\
     \mathrm{s.t.} \; t = Du, s=v
     \end{array}}} \mu \sum_{i=1}^N \phi(||t_i ||_{2}; a) + ||\mathcal{P} s ||_* + i_{\mathcal{C}}(u,v)
\label{eq:TV-LPR_ADMM}
\end{equation}
where the constraint $f = \mathcal{M}(u+v)$ is imposed by adding $i_{\mathcal{C}}$, the indicator function of the convex set $\mathcal{C}:= \{x:=(u,v) \in \mathbb{R}^{2N}\;  \mathrm{s.t.} \;  f = \mathcal{M}(u+v)\}$,  to the cost function in \eqref{eq:TV-LPR}. We propose the following Alternating Direction 
Method of Multipliers iterative scheme, where we denote by $\rho_t$ and $\rho_s$ the ADMM penalty parameters, by $x:= \left( \begin{array}{c}
     u  \\
     v 
\end{array}\right)$ and $y:= \left(\begin{array}{c}
     y_t  \\
     y_s  
\end{array}\right)$ the vector of Lagrange multipliers where $y_t \in \mathbb{R}^{2N}$ and $ y_s \in \mathbb{R}^N$. Given the initial guesses $x^0 := (u^0, v^0), y^0 := (y_t^0, y_s^0)$, the $k$-th iteration reads as
\begin{eqnarray}
    t^{k} &\in&\argmin{t \in \mathbb{R}^{2N}} \sum_{i=1}^N \mu \phi(||t_i||_2;a) + \frac{\rho_t}{2} \left\|t- \left(\mathrm{D}u^{k-1} + \frac{y_t^{k-1}}{\rho_t} \right)\right\|^2 \label{t_subpb}\\
    s^{k} &\in& \argmin{s \in \mathbb{R}^N}  \left\|\mathcal{P}s\right\|_* + \frac{\rho_s}{2} \left\|s-\left(v^{k-1}+\frac{y_s^{k-1}}{\rho_s}\right)\right\|^2 \label{s_subpb}\\
    x^{k} &\in& \argmin{x\;\in \;\mathcal{C}} \frac{\rho_t}{2} \left\| t^{k} - \frac{y_t^{k-1}}{\rho_t} - \left[ \begin{array}{cc}
     \mathrm{D}    & \mathrm{0}_N  \\
    \end{array}\right] x\right\|^2 + \frac{\rho_s}{2} \left\|s^{k}-\frac{y_s^{k-1}}{\rho_s} - \left[ \begin{array}{cc}
     \mathrm{0}_N    & \mathrm{I}_N  \\
    \end{array}\right]x\right\|^2 \label{uv_subpb} \\
    y^{k} &=& y^{k-1} + 
    \left[ \begin{array}{c}
       \rho_t \mathrm{D}\\
         \rho_s \mathrm{I}_N 
    \end{array} \right] x^k  -
    \left( \begin{array}{c}
         \rho_t t^k  \\
          \rho_s s^k
    \end{array} \right)
    \label{eq:y_subpb}
\end{eqnarray}
 \\ In the following, we describe the numerical solution of problems \eqref{t_subpb}-\eqref{uv_subpb}.
\\
\noindent\textbf{$t$-subproblem, estimation of gradients}: The minimization problem  \eqref{t_subpb} is separable with respect to $t_i=(\mathrm{D}u)_i \in \mathbb{R}^2$ hence is equivalent to $N$ optimization problem of the form:
\begin{equation}
    \min_{t_i \in \mathbb{R}^2} \mu \phi(||t_i||_2; a) + \frac{\rho_t}{2} \left\| t_i - \left( (\mathrm{D}u^{k-1})_i + \frac{(y_t^{k-1})_i}{\rho_t}\right) \right\|^2.
    \label{eq:ti_closedform}
\end{equation}
Sufficient conditions to ensure that problem \eqref{eq:ti_closedform} has a unique solution are presented in Proposition~\ref{prop1}. For a detailed proof, see \cite{Non-convexPHI}.
\begin{proposition}\label{prop1}
    Given $v \in \mathbb{R}^2$, $a \geq 0$, $\mu, \rho_t > 0$, the function $\frac{\mu}{\rho_t}\phi(\cdot;a) + \frac{1}{2}||t_i - v||^2$ is strongly convex if $a \leq \frac{\rho_t}{\mu}$. Morever, it has a unique global minimizer that has the following closed-form expression:
    \begin{equation}
        t_i = \left\{ \begin{array}{l}
           \min \left(1, \frac{1}{1-a\frac{\mu}{\rho_t}}\max\left(1-\frac{\mu}{\rho_t ||v||_2},0\right)\right) v \; \hspace{0.1cm} \mathrm{if}\; ||v||_2 > 0   \\
         0  \quad \quad \quad \qquad \qquad \qquad \qquad \qquad \qquad \; \; \;  \mathrm{if} \;||v||_2 = 0
        \end{array} \right.
        \label{eq:PHITV}
    \end{equation}
\end{proposition}
Equation \eqref{eq:PHITV} is a generalization of the proximal operator of the $\ell_2$-norm (recovered for $a \to 0$) where a second shrinkage value, that depends on $a$, is introduced. \\
\noindent\textbf{$s$-subproblem, estimation of LR texture}: The solution of the minimization problem in \eqref{s_subpb} is the evaluation of the proximal operator of the nuclear norm in $v^{k-1}+ \frac{y_s^{k-1}}{\rho_s}$. This proximal operator has the following closed form solution:
\begin{equation}
  \mathrm{prox}_{||\cdot||_*, \beta}(x) = U \mathrm{max}(\mathrm{S} - \beta \mathrm{I}) V^T  
  \label{eq:SVT}
\end{equation}
where $x = \mathrm{U} \mathrm{S} \mathrm{V^T}$
is the Singular Value Decomposition (SVD) of $x$
and the maximum is taken elementwise. \\
\noindent\textbf{$x$-subproblem, enforcing the constraint}:
The minimization problem in \eqref{uv_subpb} is a linearly constrained quadratic programming that does not admit a general closed form solution but which can be efficiently solved by means of the Projected Gradient Descent.
The cost function is differentiable with respect to $x$ and the projection onto the set $\mathcal{C}$ can be computed as in \cite{PnP_joint}. Starting from $x_0$, the algorithm reads as
\begin{equation}
   x_{i+1} = \mathcal{P}_{\mathcal{C}}\left(x_i - \tau \left(\mathrm{A} x_i - \mathrm{B} z \right) \right), \; i=0,\dots, n,
    \nonumber
    \label{eq:projGD}
\end{equation}
where $\tau$ is the gradient-step parameter, $\mathcal{P}_{\mathcal{C}}$ denotes the projection onto the set $\mathcal{C}$ and
\begin{equation}
    \mathrm{A} = \left[\begin{array}{cc}
      \rho_t \mathrm{D}^{\mathrm{T}} \mathrm{D}  & \mathrm{O} \\
      \mathrm{O} & \rho_s \mathrm{I}
    \end{array} \right], \;  \mathrm{B} = \left[ \begin{array}{cc}
      \rho_t \mathrm{D}  & \mathrm{O} \\
      \mathrm{O} & \rho_s \mathrm{I}
    \end{array} \right] \; \mathrm{and} \; z = \left(
    \begin{array}{c}
         t^{k} - \frac{y_t^{k-1}}{\rho_t} \\
          s^{k} - \frac{y_s^{k-1}}{\rho_s}
    \end{array}
    \right).
\end{equation}
We remark that the proposed ADMM algorithm has no guarantee of convergence, except for $a=0$, when the cost function is convex. For the general case, as detailed in \cite{ADMMconvergence}, a Lipschitz-gradient term is required. This would imply the replacing of the strong constraint by some $L^2$ penalization, that could also handle noisy cases.
Nevertheless, we prefer to hardly force $u$ and $v$ to satisfy the constraint.
\section{LPR-NET architecture}
Starting from the ADMM iterative scheme proposed in \eqref{t_subpb}-\eqref{eq:y_subpb}, we propose an unrolled neural network by stacking $\mathrm{K}$ blocks, each one resembling one iteration of the iterative algorithm considered. As described in Section 2, one iteration of the proposed ADMM consists of four subproblems to be solved, corresponding to the $t$, $s$, $x$ and $y$ updates. Hence, the generic $k$-th block of our network will be divided into four consecutive subblocks reproducing respectively the four subproblems. In Figure 2, the flow of a generic block of the network is shown. \\
\begin{figure}[H]
\centering
\begin{tikzpicture}
 \draw (0.45,0.45) rectangle (1.45,1.45) node at (0.95,0.95) {$f$}; 
 \draw[thick] (1.45,0.95) -- (8.75,0.95);
 \draw[thick,-stealth] (2.9,0.95) -- (2.9,-0.25);
 \draw[thick,-stealth] (4.85,0.95) -- (4.85,-0.25);
 \draw[thick,-stealth] (6.8,0.95) -- (6.8,-0.25);
 \draw[thick,-stealth] (8.75,0.95) -- (8.75,-0.25);
 \draw (0.45,-0.25) rectangle (1.45,-1.25) node at (0.95,-0.75) {$x^{k-1}$}; 
 \draw (0.45,-1.5) rectangle (1.45,-2.5) node at (0.95,-2) {$y^{k-1}$}; 
 \draw[thick,-stealth] (1.45,-0.75) -- (2.3,-0.75);
 \draw[thick,-stealth] (1.45,-2) -- (2.3,-2);
 \draw (2.3,-0.25) rectangle (3.5,-2.5) node at (2.9,-1.2){$\mathcal{T}_{k,\Theta_{k,t}}$} node at (2.9,-1.6) {\eqref{t_subpb}};
 \draw[thick,-stealth] (3.5,-1.35) -- (4.25,-1.35) 
node at (3.85,-1.15) {$t^k$};
 \draw (4.25,-0.25) rectangle (5.45,-2.5) node at (4.85,-1.2){$\mathcal{S}_{k,\Theta_{k,s}}$} node at (4.85,-1.6) {\eqref{s_subpb}};
  \draw[thick,-stealth] (5.45,-1.35) -- (6.2,-1.35) node at (5.8,-1.15) {$s^k$};
   \draw (6.2,-0.25) rectangle (7.4,-2.5) node at (6.8,-1.2){$\mathcal{X}_{k,\Theta_{k,x}}$} node at (6.8,-1.6) {\eqref{uv_subpb}};
 \draw[thick,-stealth] (7.4,-1.35) -- (8.15,-1.35) node at (7.75,-1.15) {$x^k$};
   \draw (8.15,-0.25) rectangle (9.35,-2.5) node at (8.75,-1.2){$\mathcal{Y}_{k,\Theta_{k,y}}$} node at (8.75,-1.6) {\eqref{eq:y_subpb}};
   \draw[thick,-stealth] (9.35,-0.75) -- (10,-0.75);
  \draw[thick,-stealth] (9.35,-2) -- (10,-2);
  \draw (10,-0.25) rectangle (11,-1.25) node at (10.5,-0.75) {$x^{k}$}; 
 \draw (10,-1.5) rectangle (11,-2.5) node at (10.5,-2) {$y^{k}$}; 
\end{tikzpicture}
\caption{Flow of the $k$-th block of the network. Each operation performed in a block corresponds to a step of the ADMM algorithm described in  \eqref{t_subpb}-\eqref{eq:y_subpb}. }
\end{figure}
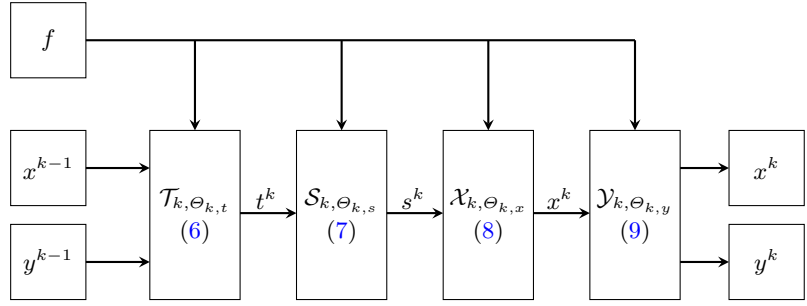
We define the output of the  $k$-th block as
\begin{equation}
    (x^{k}, y^{k}) = \mathcal{L}_{\Theta_k}^k(x^{k-1}, y^{k-1}) := \mathcal{Y}_{\Theta_{k,y}} \circ \mathcal{X}_{\Theta_{k,x}} \circ \mathcal{S}_{\Theta_{k,s}} \circ 
    \mathcal{T}_{\Theta_{k,t}}(x^{k-1},y^{k-1})
\end{equation}
where $\Theta_{k,t}$, $\Theta_{k,s}$, 
$\Theta_{k,x}$ and $\Theta_{k,y}$ 
denote the set of learnable parameters in $\mathcal{T}_{k,\Theta_{k,t}}$, $\mathcal{S}_{k,\Theta_{k,s}}$, $\mathcal{X}_{k,\Theta_{k,x}}$, $\mathcal{Y}_{k,\Theta_{k,y}}$ subblocks respectively. 
Finally, composing $\mathrm{K}$ blocks, we have the final relation between the input and the output of the network:
\begin{equation}
    (x^K, y^K) = \mathcal{F}_{\Theta}(x^0, y^0) = \mathcal{L}_{\Theta_K}^K \circ \dots \circ \mathcal{L}_{\Theta_1}^1 ( x^0, y^0),
\end{equation}
where $\Theta_k = \{\Theta_{k,t}, \Theta_{k,s}, \Theta_{k,x}, \Theta_{k,y}\}$ and $\Theta:= \cup_{k=1}^\mathrm{K} \Theta_k$ denotes the set of the learnable parameters of the network $\mathcal{F}_\Theta$. Various design options can be explored when constructing the subblocks $\mathcal{T}_{k,\Theta_{k,t}}$, $\mathcal{S}_{k,\Theta_{k,s}}$, $\mathcal{X}_{k,\Theta_{k,x}}$, $\mathcal{Y}_{k,\Theta_{k,y}}$ depending on the desired level of flexibility and complexity of $\mathcal{F}_{\Theta}$. \\
$\mathbf{\mathcal{T}_{k,\Theta_{k,t}}}$-\textbf{subblock, estimation of gradients}: The closed form expression of \eqref{eq:PHITV} can be recast as a linear layer followed by a non linear function as follows:

\begin{equation}
    t^{k} = \eta_1\left(\mathrm{D}^{(k)} u^{k-1} + \frac{y_t^{k-1}}{\rho_t}; \mu^{(k)}, \rho_t^{(k)}\right) 
\end{equation}
where $\eta_1$ is the non linear function defined by the proximal operator of the function $\phi(\cdot;a)$ defined in \eqref{eq:PHITV}. 
We replace the discrete operator $\mathrm{D}$ with a convolutional operator $\mathrm{D}_{\mathcal{T}}^{(k)}$ with kernel size $2 \times 2$ and 2 output channel. We also learn  the model parameter $\mu^{(k)}$ that is free to vary in different blocks while the penalty parameter $\rho_t$ is shared across them. To ensure the condition in Proposition 1 is satisfied, we parametrize $a^{(k)}$ as $
    a^{(k)} = \frac{\rho_t}{\mu^{(k)}} \left(1 - \frac{1}{b^{(k)}}\right) $
where $b^{(k)}$ is a learnable parameter constrained to be larger than 1. \\
$\mathbf{\mathcal{S}_{k,\Theta_{k,s}}}$-\textbf{subblock, estimation of LR texture}: As described in Equation \eqref{eq:SVT}, updating $s$ involves the computation of the SVD decomposition of the input since we need to apply a threshold on its singular values. In this work, we do not explore possible linear approximations of the SVD decomposition and we compute it exactly. Therefore, we preserve the same structure of \eqref{eq:SVT} adding flexibility by learning a shared $\rho_s$. Finally, the patch operator requires the size of the patch $p$ and the overlap $o$ between adjacent patches and a good choice is to fix $p=4$ and $o=2$. \\
$\mathbf{\mathcal{X}_{k,\Theta_{k,x}}}$-\textbf{subblock,  enforcing the constraint}: The algorithm described in \eqref{eq:projGD} can be viewed as a composition of $n$ layers, each one made of a linear layer followed by a non linear function, as illustrated in details in Figure 3. To preserve the specific form of the algorithm while gaining flexibility, we replace the discrete gradient operator $\mathrm{D}$ and its transpose $\mathrm{D}^{\mathrm{T}}$ with two convolutional operators with kernel size $2 \times 2$ and 2 and 1 output channels respectively. We denote them with $\mathrm{D}_{\mathcal{X}}^{(k)}$ and $\mathrm{\tilde{D}}_{\mathcal{X}}^{(k)}$ allowing them to change in each block $\mathcal{X}_{k,\Theta_{k,x}}$ and relaxing the matching between $\mathrm{\tilde{D}}_{\mathcal{X}}^{(k)}$ and $(\mathrm{D}_{\mathcal{X}}^{(k)})^{\mathrm{T}}$. Finally, the parameter $\tau$ is learned and shared across different blocks.
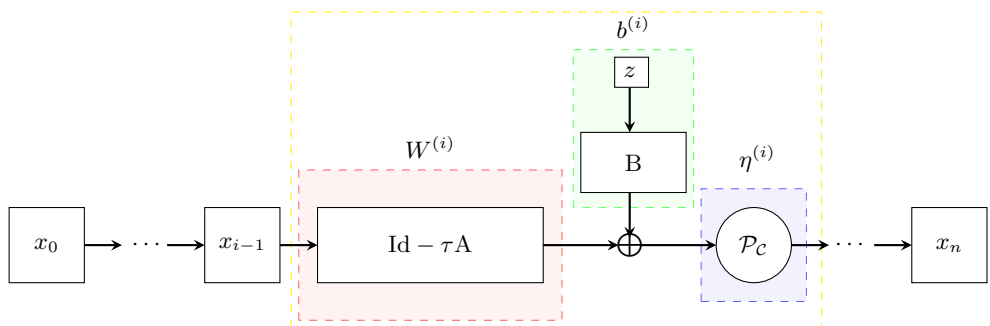
\begin{figure}[H]
\centering
\begin{tikzpicture}
    \def\a{2.6}
    \draw (0,0) rectangle (1,1) node at (0.5,0.5) {$x_0$};
    \draw[thick,-stealth] (1,0.5) -- (1.5,0.5);
    \node at (1.85,0.5) {$\cdots$};
     \draw[thick,-stealth] (2.1,0.5) -- (2.6,0.5);
     \draw (2.6,0) rectangle (3.6,1) node at (3.1,0.5) {$x_{i-1}$};
    \begin{scope}[on background layer]
    \filldraw[color=red!60, fill=red!5, dashed] (1.25+\a,-0.5) rectangle (4.75+\a,1.5);
  \end{scope}
  \begin{scope}[on background layer]
    \draw[color=yellow, dashed] (1.15+\a,-0.6) rectangle (8.2+\a,3.6);
  \end{scope}
   \node at (3+\a,1.8) {$W^{(i)}$};
    \draw[thick,-stealth] (1+\a,0.5) -- (1.5+\a,0.5);
    \filldraw[color = black, fill = white] (1.5+\a,0) rectangle (4.5+\a,1) node at (3+\a,0.5) {$\mathrm{Id} - \tau \mathrm{A} $};
    \draw[thick,-stealth] (4.5+\a,0.5) -- (5.5+\a,0.5);
    \node at (5.65+\a,0.5) {$\bigoplus$};
    \draw[thick,-stealth] (5.8+\a,0.5) -- (6.8+\a,0.5);
  \filldraw[color=black, fill = white] (7.3+\a,0.5) circle (0.5 cm) node at (7.3+\a,0.5) {$\mathcal{P}_{\mathcal{C}}$};
    \begin{scope}[on background layer]
    \filldraw[color=blue!60, fill=blue!5, dashed] (6.6+\a,-0.25) rectangle (8+\a,1.25);
  \end{scope}
   \node at (7.35+\a,1.65) {$\eta^{(i)}$};
    \draw[thick,-stealth] (5.65+\a,1.2) -- (5.65+\a,0.6);
    \filldraw[color=black, fill=white] (5+\a,2) rectangle (6.4+\a,1.2) node at (5.7+\a,1.6) {$\mathrm{B}$};
    \draw[thick,-stealth] (5.65+\a,2.6) -- (5.65+\a,2);
    \filldraw[color=black, fill=white] (5.45+\a,3) rectangle (5.9+\a,2.6) node at (5.65+\a,2.8) {$z$};
    \begin{scope}[on background layer]
    \filldraw[color=green!60, fill=green!5, dashed] (4.9+\a,1) rectangle (6.5+\a,3.1);
  \end{scope}
  \node at (5.7+\a,3.4) {$b^{(i)}$};
    \draw[thick,-stealth] (7.8+\a,0.5) -- (8.3+\a,0.5);
    \node at (8.6+\a,0.5) {$\cdots$};
   \draw[thick,-stealth] (8.9+\a,0.5) -- (9.4+\a,0.5);
   \draw (9.4+\a,0) rectangle (10.4+\a,1) node at (9.9+\a,0.5) {$x_{n}$};
\end{tikzpicture}
\caption{Unrolling of the Projected Gradient Descent algorithm.}
\label{fig:PGD_unrolled}
\end{figure}
\hspace{-0.5cm}$\mathbf{\mathcal{Y}_{k,\Theta_{k,s}}}$-\textbf{subblock}: in the updating of the Lagrange multipliers, we only replace the discrete gradient operator $\mathrm{D}$ with a convolutional kernel of size $2 \times 2$ with two channel denoted as $\mathrm{D}_{\mathcal{Y}}^{(k)}$.\\
Specifically, we evaluate and compare four distinct configurations of parameters with the twofold aim of studying the improvement gained with:
\begin{itemize}
    \item[$\bullet$] a non-convex penalty $\phi$ (a>0) with respect to the Total variation (a=0);
    \item[$\bullet$] a data-dependent $\mu^{(k)}$, estimated on $x^k$ after each block $\mathcal{L}_{\Theta_k}^k$ with respect to a fixed $\mu^{(k)}$. In the former, the estimation is performed by adding a small CNN composed of three convolutional layers with kernel size $3 \times 3$ and 4,8,16  channels respectively, each one followed by a ReLU, and in the end, we have a fully connected layer and a Softplus to enforce the positivity of the estimated parameter $\mu^{(k)}$.
\end{itemize}
By combining a convex ($a=0$) or non-convex total variation ($a>0$) and a data-dependent or fixed $\mu^{(k)}$, we obtain four configurations. For each configuration, the learnable parameters are summarized as follows in Table 1:
\begin{table}
    \centering
    \begin{tabular}{|c|c|c|c|c|c|}
    \hline
     &  $\Theta_{k,t}$   &  $\Theta_{k,s}$ &  $\Theta_{k,x}$ & $\Theta_{k,y}$ & $|\Theta|$\\
     \hline
      $\Theta_1$  & $a^{(k)}= 0, \mu^{(k)}, \rho_t, \mathrm{D}_{\mathcal{T}}^{(k)}$ & $ \rho_s $ &    $\mathrm{D}_{\mathcal{X}}^{(k)} , \tilde{\mathrm{D}}_{\mathcal{X}}^{(k)}, \tau$ & $\mathrm{D}_{\mathcal{Y}}^{(k)}$ & $\approx 50 \mathrm{K}$ \\
      \hline
      $\Theta_2$  & $b^{(k)}, \mu^{(k)}, \rho_t, \mathrm{D}_{\mathcal{T}}^{(k)}$ & $ \rho_s $ &    $\mathrm{D}_{\mathcal{X}}^{(k)} , \tilde{\mathrm{D}}_{\mathcal{X}}^{(k)}, \tau$ & $\mathrm{D}_{\mathcal{Y}}^{(k)}$ & $\approx 50 \mathrm{K}$ \\
      \hline
      $\Theta_3$  & $a^{(k)}=0, \mathrm{CNN}^{(k)} , \rho_t, \mathrm{D}_{\mathcal{T}}^{(k)}$ & $ \rho_s $ &    $\mathrm{D}_{\mathcal{X}}^{(k)} , \tilde{\mathrm{D}}_{\mathcal{X}}^{(k)}, \tau$ & $\mathrm{D}_{\mathcal{Y}}^{(k)}$ & $\approx 2 \cdot 10^3 \mathrm{K}$ \\
      \hline 
      $\Theta_4$  & $b^{(k)}, \mathrm{CNN}^{(k)} , \rho_t, \mathrm{D}_{\mathcal{T}}^{(k)}$ & $ \rho_s $ &    $\mathrm{D}_{\mathcal{X}}^{(k)} , \tilde{\mathrm{D}}_{\mathcal{X}}^{(k)}, \tau$ & $\mathrm{D}_{\mathcal{Y}}^{(k)}$ & $\approx 2 \cdot 10^3 \mathrm{K}$\\
      \hline 
    \end{tabular}
    \vspace{0.1cm}
    \caption{Four different parameter configurations with an estimate of the number of parameters.}
    \label{tab:configurations}
\end{table}
Configurations $\Theta_1$ and $\Theta_2$ contain less parameters and a lower representation power than $\Theta_3$ and $\Theta_4$, whereas configurations $\Theta_2$ and $\Theta_4$ should take advantage of the non-convex setting to better recover piecewise constant images. A detailed analysis will be carried out in Section 4.

\section{Numerical Experiments}
In this section, the training procedure of our LPR-NET is explained in detail and some numerical results are presented. In the first example, we will investigate the role that $\mathrm{K}, n$ and the four different configurations play in the final output results while in the second example, we present decomposition of synthetic and natural images compared with state-of-the-art methods.\\
\textbf{Training strategy}.
We generate a synthetic image dataset composed of 400 images of dimension $64 \times 64$ that are the sum of a piecewise constant image $u^{GT}$ and a texture image $v^{GT}$ eventually degraded by a mask operator $\mathcal{M}$. The structure component is obtained by randomly positioning connected supports generated via the Lane-Riesenfeld algorithm while sparse Fourier texture generation has been used for the texture component. For further details, see \cite{PnP_joint}. 
Given our synthetic dataset and our architecture $\mathcal{F}_{\Theta}$, the estimate of the set of parameters $\Theta$ is carried out via the minimization of the Mean Square Error over the training dataset:
\begin{equation}
    \widehat{\Theta} \in \argmin{\Theta}
    \frac{1}{N_{train}} \sum_{j=1}^{N_{train}} \left\| \mathcal{F}_{\Theta}(x^0_j,y^0_j) - x_j^{\mathrm{GT}} \right\|^2,
\end{equation}
where $x_j^0=(u_j^0,v_j^0)$ and $y_j^0$ are the initial guesses of the ADMM set as $u_j^0 = f$, $v_j^0 = 0$ and $y_j^0 = 0$. At the $k$-th block of the network, the initial guess for $x_j$ in the Projected Gradient Descent is chosen as $x^{k-1}_j$. The network has been implemented on Pytorch and trained using Adam optimizer with learning rate = $10^{-3}$ halved after forty epochs, training batch size = 16 and for a total of 250 epochs. We initialized the parameters as: $(\mu^{(k)})^0 = 0.05$, $\rho_t^0 = 1 $, $\rho_s^0 = 1 $, $\tau^0 = 0.1$, $(b^{(k)})^0 = 1.1$, $\mathrm{D}_{\mathcal{T}}^{(k)}, \mathrm{D}_{\mathcal{X}}^{(k)}, \mathrm{D}_{\mathcal{Y}}^{(k)}$ initialized as the discrete gradient, $\tilde{\mathrm{D}}_{\mathcal{X}}^{(k)}$ initialized as the transpose of the discrete gradient and the parameters of the small CNN are initialized such that the output of the network is equal to $(\mu^{(k)})^0 = 0.05$ (all zeros except the bias of the last fully connected layer).  \\
\textbf{Example 1: Ablation study of the LPR-NET}.
In this example, we trained our LPR-NET for different values of the number of unrolled iterations of the ADMM $\mathrm{K}$ and of the Projected Gradient Descent $n$ over the training dataset of 400 images considering no degradation $\mathcal{M}$. In particular, we chose $\mathrm{K}=5,10,15,20$ and $n= 5,10,15,20$ and quantitatively compared the different networks on a test dataset of 1000 images generated from the synthetic generator of structure and texture images available at \cite{joint_git}. In Table 2, the mean PSNR values of the dataset are reported for the four configurations of parameters. 

\begin{table}
    \centering
    \begin{tabular}{ccccccccccc|cccccccccccc}
    \cline{1-23}
 \multicolumn{11}{c|}{$\Theta_1$} & &
 \multicolumn{11}{c}{$\Theta_2$} \\
    \cline{1-23}
    & \multicolumn{3}{c}{$n$} & 5 & & 10 & &15 & & 20 & & & \multicolumn{3}{c}{$n$} & 5 & & 10 & &15 & & 20 \\
    \cline{1-23}
     \multirow{4}{*}{K} & & 5 & & 38.91 & & 39.03 & & 39.11 & & 39.23 & & \multirow{4}{*}{K} & & 5 & & 39.14 & & 39.13 & & 38.93 & & 39.09 \\
     \cline{2-11}
     \cline{14-23}
      & & 10 &  & 40.44 & &  40.38 & & 40.48 &  & 40.49 & & & & 10 &  & 40.78 & &  40.61 & & 40.60 &  & 40.61  \\
      \cline{2-11}
      \cline{14-23}
      & & 15 & & 41.08 & & 41.11 & & 41.01  & & 40.94 & & & & 15 & & 41.44 & & 41.34 & & 41.28  & & 41.24 \\
      \cline{2-11}
      \cline{14-23}
      & & 20 & & 41.60 & & 41.44 & & 41.35  & & 41.38  & & & & 20 & & 42.00 & & 41.75 & & 41.68  & & 41.74 \\
      \cline{1-23} 
      \multicolumn{11}{c|}{$\Theta_3$} & &
 \multicolumn{11}{c}{$\Theta_4$} \\
    \cline{1-23}
    & \multicolumn{3}{c}{$n$} & 5 & & 10 & &15 & & 20 & & & \multicolumn{3}{c}{$n$} & 5 & & 10 & &15 & & 20 \\
    \cline{1-23}
     \multirow{4}{*}{K} & & 5 & & 39.23 & & 39.34 & & 39.38 & & 39.40 & & \multirow{4}{*}{K} & & 5 & & 39.52 & & 39.51 & & 39.55 & & 39.54 \\
     \cline{2-11}
     \cline{14-23}
      & & 10 &  & 40.64 & &  40.49 & &40.41 &  & 40.39 & & & & 10 &  & 41.02 & &  40.93 & & 40.82 &  & 40.93  \\
      \cline{2-11}
      \cline{14-23}
      & & 15 & & 41.33 & & 41.19 & & 41.13  & & 41.00 & & & & 15 & & 41.69 & & 41.69 & & 41.60  & & 41.46 \\
      \cline{2-11}
      \cline{14-23}
      & & 20 & & 41.81 & & 41.51 & & 41.43  & & 41.41  & & & & 20 & & 42.13 & & 41.96 & & 41.76  & & 41.65 \\
      \cline{1-23} 
      
    \end{tabular}
    \caption{mean PSNR over the test dataset for different values of $\mathrm{K}$ and $n$ and for the four different configurations.}
    \label{tab:mean_PSNR}
\end{table}
\noindent As expected, increasing the number of blocks of the network $\mathrm{K}$ leads to better quality decomposition results while higher values of $n$ show a small deterioration.
Among the four configurations, adding a non-convex penalty $\phi$ improved the decomposition with respect to $a=0$ while adding a small CNN to make $\mu^{(k)}$ data-dependent brings a slight improvement when compared to a variable $\mu^{(k)}$ shared for all images. Note that we limit our ablation study to $\mathrm{K}$ and $n$ but could be extended to the size of filters, on matching and untied adjoints and that more flexibility could be achieved by allowing larger filters and more channels. \\ 
\textbf{Example 2: comparison with state-of-the-art methods}.
In this example, we compare the best configuration of our LPR-NET with three variational models - TV-$L_2$, TV-G, LPR - and with the joint structure-texture decomposition model $R_{x}(u,v)$ proposed in \cite{PnP_joint}, reporting the mean PSNR over the same test dataset in Table 3. 
\begin{table}
    \centering
    \begin{tabular}{cccccc}
        \hline
      & $\Theta_4$   & TV-$L_2$ \cite{TV-L2} & TV-G \cite{TV-G} & LPR \cite{LPR} & $R_x(u,v) $ \cite{PnP}\\
     \hline
      PSNR &  42.13 & 38.84  &  38.86  & 41.61   & 42.69 \\
    \hline
    \end{tabular}
    \caption{Comparison of mean PSNR values with classical variational model and \cite{PnP_joint}. Notice that the regularization parameter $\mu$ was set by trials and errors for the 3 variational models to achieve the best possible PSNR.}
    \label{tab:comp_others}
\end{table}
The best configuration of our LPR-NET outperforms classical variational models whose parameters are selected in order to get the best PSNR values for each distinct image. This means that, despite, on average, our results are worse than the $R_{x}(u,v)$ model, our network is able to automatically set the optimal parameters adapting to the image content avoiding having to set-up the parameters of the Plug-and-Play optimization algorithm. We also remark that, with respect to \cite{PnP_joint}, our network has a lighter architecture (around $4 \times 10^4$ parameters) that requires a smaller training dataset and less training hours (in the worst case, around 4 hours). In Figure 4 and Figure 5, we illustrate decomposition results of synthetic and natural images compared with $R_{x}(u,v)$.   
\begin{figure}
    \centering
    \begin{tabular}{c|ccc}
\includegraphics[width=0.2\linewidth]{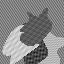} &   \includegraphics[width=0.2\linewidth]{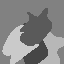} & \includegraphics[width=0.2\linewidth]{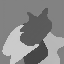} & \includegraphics[width=0.2\linewidth]{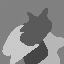} \\
$f$ & $u^{GT}$ & LPR-NET &  $R_x(u,v)$  \\
         &  \includegraphics[width=0.2\linewidth]{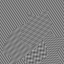} &   \includegraphics[width=0.2\linewidth]{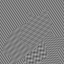} 
         &  \includegraphics[width=0.2\linewidth]{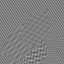}  \\    
         & $v^{GT}$ & LPR-NET &  $R_x(u,v)$ \\
    \end{tabular}
    \caption{Decomposition results of a synthetic image obtained with our LPR-NET with $(\mathrm{K}, n)=(20,5)$ and with the joint structure-texture model proposed in \cite{PnP_joint}(PSNR($u_{\mathrm{LPR-NET}},u^{GT}$) = 41.93, PSNR($u_{R_{x}(u,v)},u^{GT}$) = 40.53, PSNR($v_{\mathrm{LPR-NET}},v^{GT}$) = 41.80, PSNR($v_{R_{x}(u,v)},v^{GT}$) = 40.25).}
    \label{fig:synthetic}
\end{figure}
\begin{figure}
    \centering
    \begin{tabular}{c|cc}
   \includegraphics[width=0.25\linewidth]{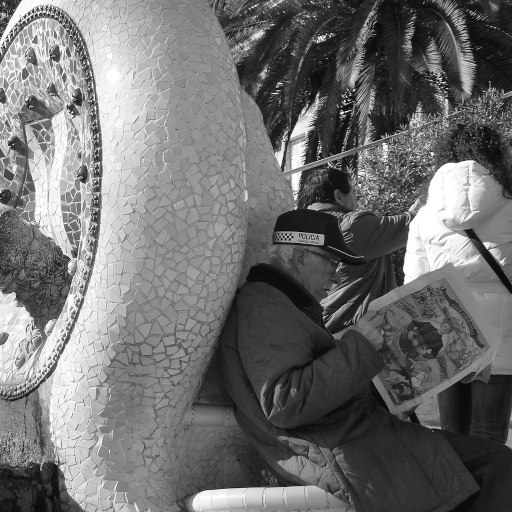}      &  \includegraphics[width=0.25\linewidth]{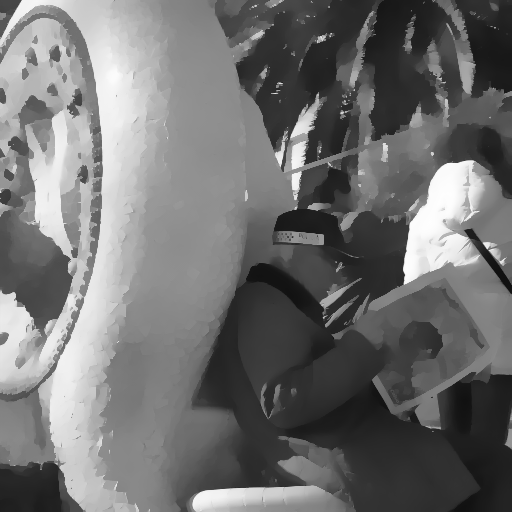} & \includegraphics[width=0.25\linewidth]{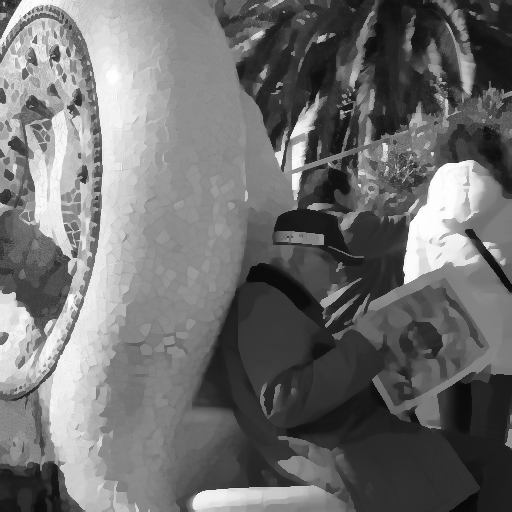} \\
  $f$ & $u$ - LPR-NET & $u$ - $R_x(u,v)$ \\
   & \includegraphics[width=0.25\linewidth]{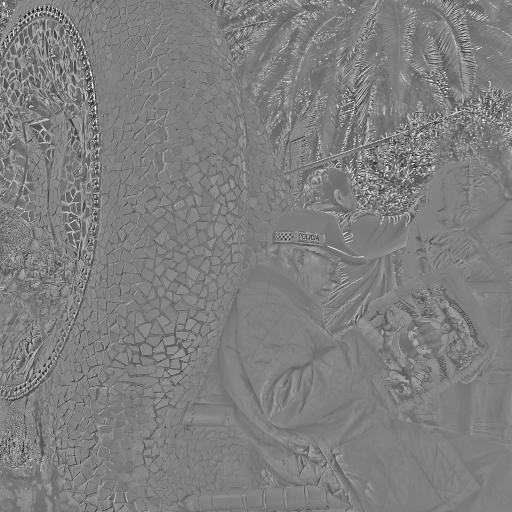} & \includegraphics[width=0.25\linewidth]{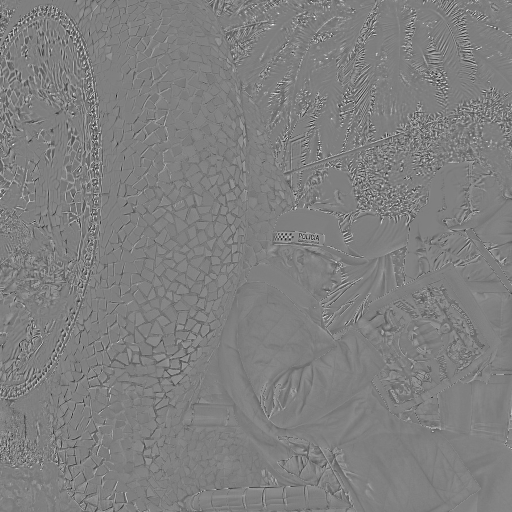} \\
   & $v$ - LPR-NET & $v$ - $R_x(u,v)$ \\
   \includegraphics[width=0.25\linewidth]{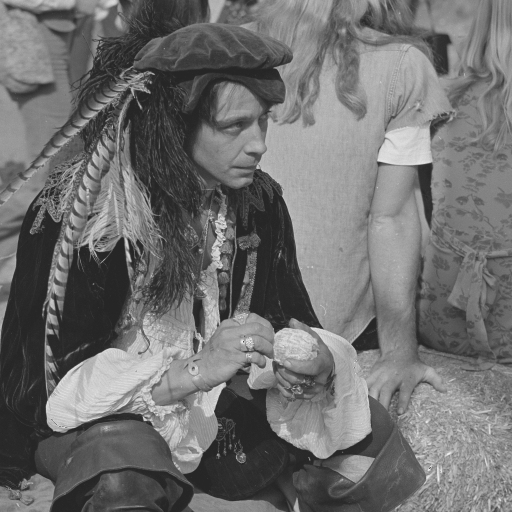}      &  \includegraphics[width=0.25\linewidth]{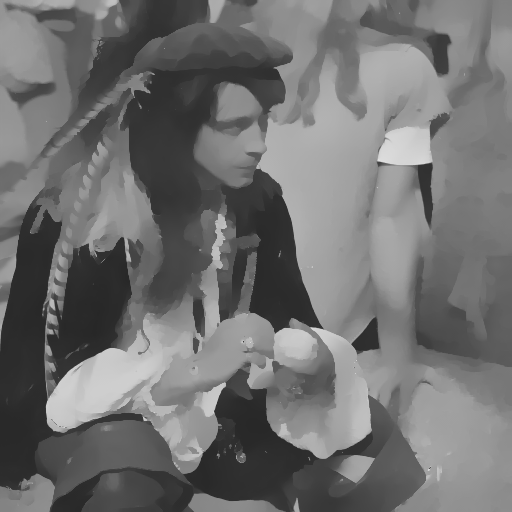} & \includegraphics[width=0.25\linewidth]{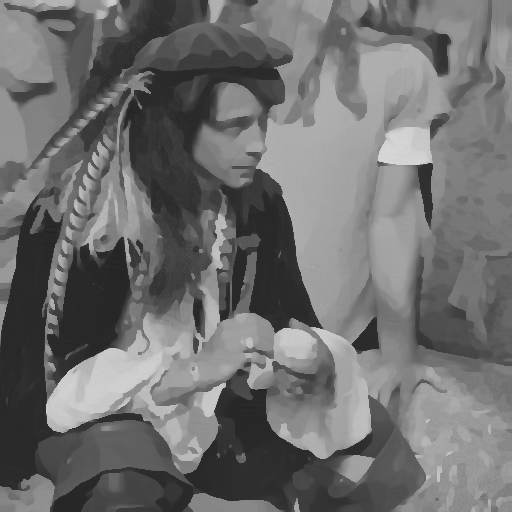} \\
   $f$ & $u$ - LPR-NET & $u$ - $R_x(u,v)$ \\
   & \includegraphics[width=0.25\linewidth]{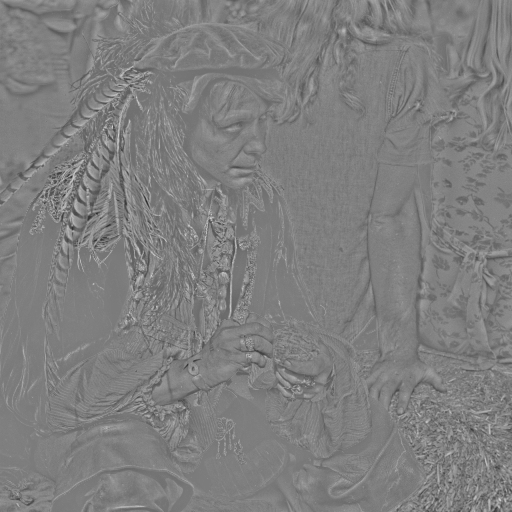} & \includegraphics[width=0.25\linewidth]{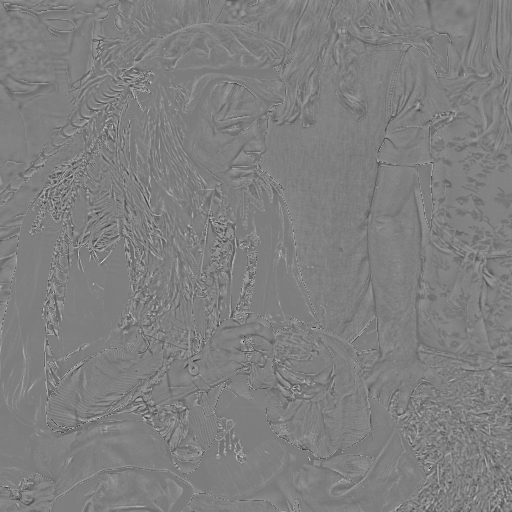} \\
   & $v$ - LPR-NET & $v$ - $R_x(u,v)$ \\
    \end{tabular}
    \caption{Decomposition results on natural images obtained with our LPR-NET with $(\mathrm{K},n)=  (20,5)$ (left) and with the model proposed in \cite{PnP_joint} (right).}
    \label{fig:natural}
\end{figure}
While for the synthetic image the results are visually comparable, we note how on natural images the behaviour is slightly different: the LPR-NET removes most of the texture but also some structural edges (e.g. the face of the pirate) while the $R_x (u,v)$ model better preserves the geometric of the image but removes few tiles in the mosaic.\\
\textbf{Extension to inverse problems}.
Finally, we also carry out some inpainting experiments. With this aim, we trained our LPR-NET (($\mathrm{K},n) = (20,20))$ following the training strategy described in Section 4 except for the training dataset where each image is masked with some random shapes. In Figure 6, we compare our model with \cite{PnP_joint} on Barbara where we randomly remove $65 \%$ of pixels. The results are promising when the measurement operator is fixed and well described by $\mathcal{M}$ as our network achieves a more "natural" decomposition with respect to the Plug-and-Play method $R_x(u,v)$  where some artifacts are visible in Barbara's arm 
(recall that $R_x(u,v)$ is trained independently of the measurement method). 
\begin{figure}
    \centering
    \begin{tabular}{c|ccc}
   \includegraphics[width=0.25\linewidth]{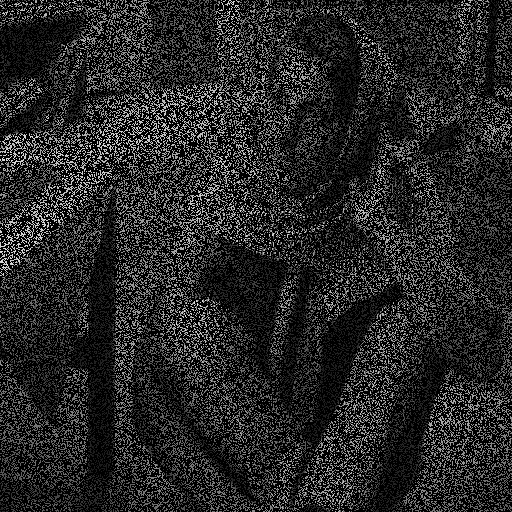}      &  \includegraphics[width=0.25\linewidth]{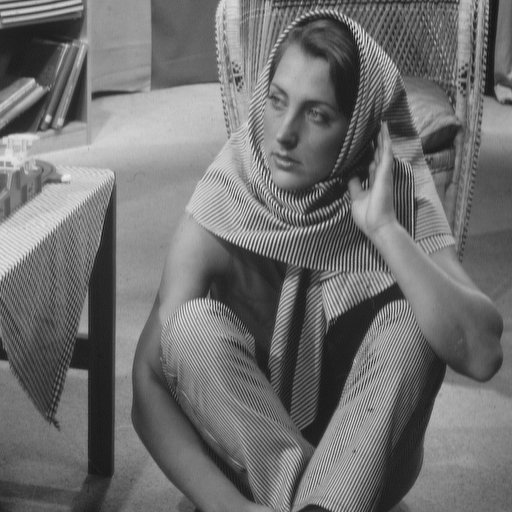} & \includegraphics[width=0.25\linewidth]{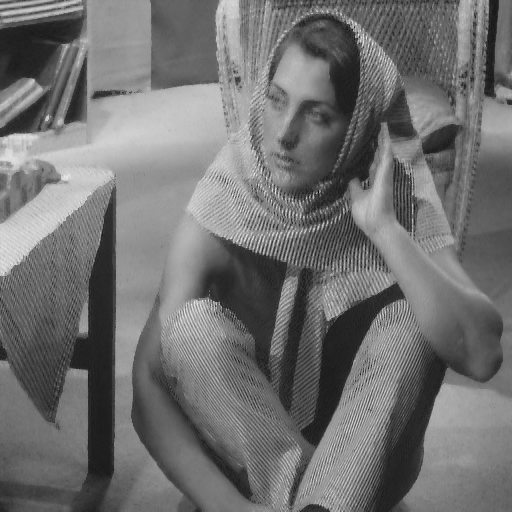} & \includegraphics[width=0.25\linewidth]{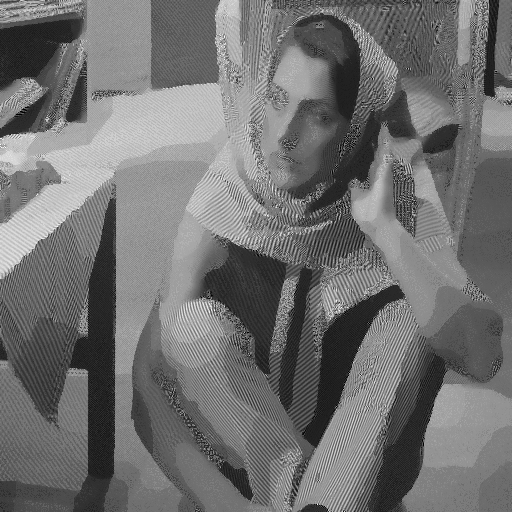} \\
   $\mathcal{M}f$ & $f$ & LPR-NET & $R_{x}(u,v)$ 
    \end{tabular}
    \caption{Inpainting results on Barbara with 65 \% of pixels randomly removed, and comparison with the model proposed in \cite{PnP_joint}.}

\end{figure}

\section{Conclusion}
In this paper, we presented a fully automatic way of separating images into structure and texture components based on the unrolling of a generalized low patch rank model. We proposed a novel unrolled network LPR-NET and investigated various architectural choices. Numerical experiments validate the potentiality of our network to achieve high quality decomposition results without tuning parameters.


\begin{thebibliography}{8}
\bibitem{OktemPD} J. Adler and O. Öktem, "Learned Primal-Dual Reconstruction", in IEEE Transactions on Medical Imaging, vol. 37, no. 6, pp. 1322-1332, 2018.
\bibitem{TV-G} J-F. Aujol, G. Aubert, L. Blanc-Feraud, and A. Chambolle,  "Image decomposition into a bounded variation component and an oscillating component", Journal of Mathematical Imaging and Vision, 22
571, pp. 71–88, 2005.
\bibitem{struct} J-F. Aujol, G. Giboa, T. Chan, and S. Osher, "Structure-texture image decomposition - Modeling, Algorithms, and Parameter selection", International Journal of Computer Vision, 67, pp. 11-136, 2006.
\bibitem{TV-L2} A. Chambolle, "An algorithm for total variation minimization and applications", Journal of Mathematical
586 imaging and vision, 20, pp. 89–97, 2004.
\bibitem{SMOOTHPLUSSPARSE} S. Ducotterd, S. Neumayer and M. Unser, "Learning of Patch-Based Smooth-Plus-Sparse Models for Image Reconstruction", arXiv preprint \url{arXiv:2412.13070
}, 2024.
\bibitem{quaternary} L. Girometti, M. Huska, A. Lanza and S. Morigi, "Quaternary image decomposition with crosscorrelation-based multi-parameter selection", In: Scale Space and Variational Methods in Computer Vision: 9th International Conference, SSVM 2023, pp. 120–133, 2023.
\bibitem{GregLecun} K. Gregor and Y. LeCun, “Learning Fast Approximations of Sparse
Coding”, in Proc. Int. Conf. Machine Learning, 2010.
\bibitem{localLPR} A. Guennec, J.-F. Aujol, Y. Traonmilin, "Adaptive parameter selection for gradient-sparse
plus low patch-rank recovery: application to image
decomposition"
\bibitem{PnP_joint} A. Guennec, J-F. Aujol and Y. Traonmilin, "Joint structure-texture low dimensional
modeling for image decomposition with a plug and play framework", ffhal-04648963, 2024.
\bibitem{joint_git} A. Guennec, J-F. Aujol and Y. Traonmilin,  \url{https://github.com/aguennecjacq/joint_decomposition}, 2024.
\bibitem{GEN_MOD} A. Habring and M. Holler, "A Generative Variational Model for Inverse Problems in Imaging", SIAM Journal on Mathematics of Data Science, vol. 4, no 1, pp. 306-335, 2022.
\bibitem{Non-convexPHI} M. Huska, A. Lanza, S. Morigi, and I. Selesnick, "A convex-nonconvex variational
method for the additive decomposition of functions on surfaces", Inverse Problems,
35, 124008, 2019.
\bibitem{PnP} Y. Kim, B. Ham, M. N. Do, and K. Sohn, “Structure–texture image
decomposition using deep variational priors,” IEEE Trans. Image
Process., vol. 28, no. 6, pp. 2692–2704, 2019.
\bibitem{den_unrolling} H. T. V. Le, N. Pustelnik and M. Foare, "The faster proximal algorithm, the better unfolded deep learning architecture ? The study case of image denoising," 30th European Signal Processing Conference (EUSIPCO), Belgrade, Serbia, pp. 947-951, 2022.
\bibitem{deblur} Y. Li, M. Tofighi, J. Geng, V. Monga, and Y. C Eldar, “Efficient and
Interpretable Deep Blind Image Deblurring via Algorithm Unrolling”,
IEEE Trans. Comput. Imaging, vol. 6, pp. 666–681, Jan. 2020
\bibitem{G-norm}Y. Meyer and D. Lewis, "Oscillating Patterns in Image Processing and Nonlinear Evolution Equations:
The Fifteenth Dean Jacqueline B. Lewis Memorial Lectures", Memoirs of the American Mathematical
Society, American Mathematical Society, 2001.
\bibitem{review} V. Monga, Y. Li, and Y. C. Eldar, “Algorithm unrolling: Interpretable,
efficient deep learning for signal and image processing”, IEEE Signal
Processing Magazine, vol. 38, no. 2, pp. 18–44, 2021.


\bibitem{BNN} S. Ono, T. Miyata, and I. Yamada, “Cartoon-texture
image decomposition using blockwise low-rank texture
characterization”, IEEE Transactions on Image Processing, vol. 23, no. 3, pp. 1128–1142, March 2014.
\bibitem{ROF} L. Rudin, S. Osher, E. Fatemi. "Nonlinear total variation based noise removal
algorithms", Physica D, 60, 259–268, 1992.
\bibitem{LPR} H. Schaeffer and S. Osher, “A low patch-rank interpretation of texture”, SIAM Journal on Imaging Sciences,
vol. 6, no. 1, pp. 226–262, 2013.

\bibitem{ADMMconvergence}  A. Themelis, P. Patrinos. "Douglas–Rachford splitting and ADMM for nonconvex optimization: tight
convergence results", SIAM J. Optim. 30(1), 149–181, 2020.

\bibitem{MCP} C. H. Zhang. "Nearly unbiased variable selection under minimax concave penalty",  Ann. Stat. 38 (2),
pp. 894–942, 2010.

\bibitem{SPARSECODING} H. Zhang and V. M. Patel, "Convolutional Sparse and Low-Rank Coding-Based Image Decomposition," in IEEE Transactions on Image Processing, vol. 27, no. 5, pp. 2121-2133, 2018.

\bibitem{FUSION} Y. Zhang, M. Yang, N. Li and Z. Yu, "Analysis-synthesis dictionary pair learning and
patch saliency measure for image fusion" Signal Process. 167, pp. 107327, 2020.

\bibitem{unsupervised} F. Zhou, Q. Chen, B. Liu, and G. Qiu, “Structure and
texture-aware image decomposition via training a neural
network”, IEEE Transactions on Image Processing, vol.
29, pp. 3458–3473, 2019.


\end{thebibliography}
\end{document}